\documentclass[letterpaper,10pt,conference]{ieeeconf}
\IEEEoverridecommandlockouts
\usepackage{amsmath}
\usepackage{amssymb}
\usepackage{cite}
\usepackage{booktabs}
\usepackage{graphicx}
\usepackage{algorithm}
\usepackage{algpseudocode}
\usepackage{hyperref}
\usepackage{color}
\usepackage{colortbl}
\usepackage[normalem]{ulem}
\definecolor{refinementblue}{RGB}{38,139,230}
\definecolor{refinementred}{RGB}{230,38,38}
\definecolor{oursrow}{RGB}{180,215,245}
\definecolor{referencegreen}{RGB}{40,180,60}
\hypersetup{colorlinks=true, linkcolor=referencegreen, citecolor=black, urlcolor=black}

\title{\LARGE\bf Fast Inference-Time Policy Refinement via Action-Affine World Models}

\author{Run Wang$^{1}$, Liang Wu$^{2}$, J\'an Drgo\v{n}a$^{2}$, Yilin Mo$^{1}$\thanks{Corresponding author: Yilin Mo. $^{1}$Run Wang and Yilin Mo are with the Department of Automation, Tsinghua University, Beijing, China. {\tt\small wangrun24@mails.tsinghua.edu.cn, ylmo@tsinghua.edu.cn}. $^{2}$Liang Wu and J\'an Drgo\v{n}a are with Johns Hopkins University, Baltimore, MD 21218, USA. {\tt\small \{wliang14,jdrgona1\}@jh.edu}.}}

\begin{document}

\maketitle

\thispagestyle{empty}

\pagestyle{empty}

\begin{abstract}
    Imitation policies can fail through compounding errors, while existing refinement methods often require additional on-policy learning or expensive online planning. Our key insight is to model the response to small corrections around a policy proposal in an affine form that enables fast optimization. We refine a frozen imitation policy using an \textit{action-affine world model} (A\textsuperscript{2}WM), trained offline from perturbed policy rollouts. Conditioned on history and the proposal, the model directly predicts a nominal trajectory and a linear action response over the full horizon. With quadratic costs and linear constraints, refinement becomes a cheap quadratic program (QP), requiring only one model evaluation and one QP solve per control step. To the best of our knowledge, we are the first to combine an action-affine world model with QP-based inference-time refinement of frozen imitation policies. Across four manipulation tasks, our framework achieves the highest success rate among evaluated baselines, adding only 4.2 ms of mean refinement latency per control step, excluding base-policy sampling.
\end{abstract}

\section{Introduction}

Behavior cloning (BC) enables robots to acquire manipulation skills from expert demonstrations. Diffusion policies (DP) represent multimodal action sequences and provide useful proposals for reaching, grasping, and assembly~\cite{chi2023diffusion}. At deployment, however, execution errors can drive the robot outside the demonstrated state distribution, where errors compound and can lead to failure~\cite{ross2011dagger}. Refining a pretrained policy through feedback offers a way to recover from such deviations. \textit{The challenge is to compute effective action corrections with sufficiently low latency to enable real-time feedback.}

\begin{figure}[t]
    \centering
    \includegraphics[width=\columnwidth]{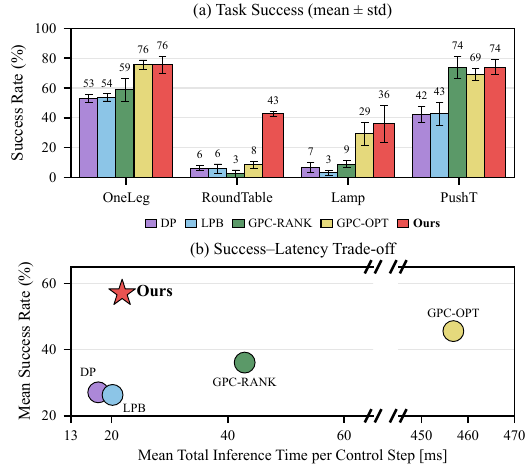}
    \caption{(a) \textbf{Task success (mean $\pm$ std across three evaluation seeds).} Ours achieves the highest observed success among the evaluated methods, tying GPC-OPT on OneLeg and GPC-RANK on PushT. (b) \textbf{Success--latency trade-off.} Ours achieves the highest four-task mean success with a modest increase in total per-control-step inference time over DP. Total inference time includes policy sampling and online refinement.}
    \label{fig:success-latency}
\end{figure}

\begin{figure*}[t]
    \centering
    \includegraphics[width=\textwidth,pagebox=cropbox,clip]{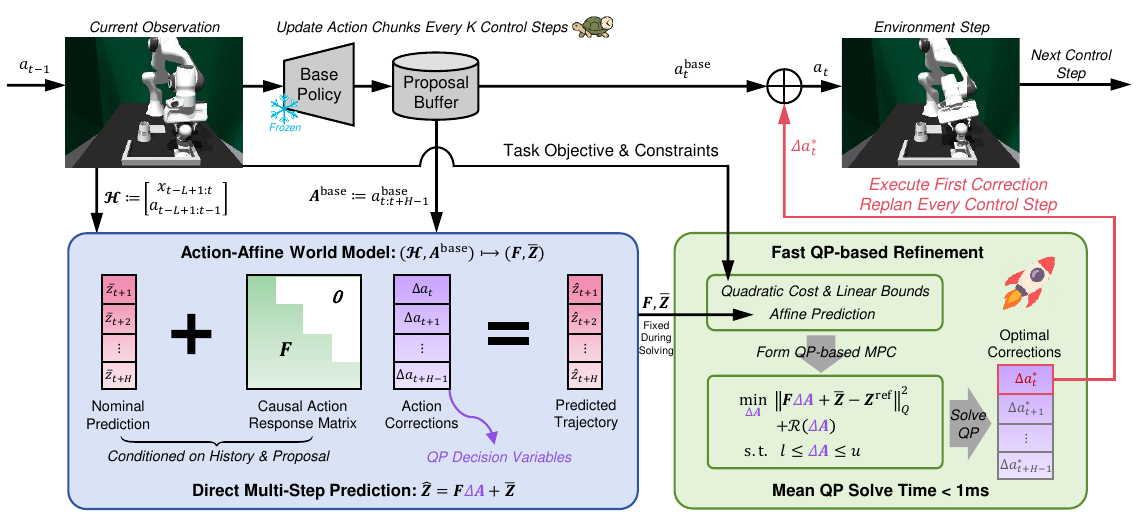}
    \caption{\textbf{Overview of fast inference-time policy refinement.} A frozen base policy generates action proposals every $K$ control steps. At each control step, the action-affine world model uses interaction history and the current proposal to predict a nominal trajectory $\bar Z$ and a causal action-response matrix $F$, yielding the direct multi-step prediction $\widehat Z=F\Delta A+\bar Z$. With $F$ and $\bar Z$ fixed, quadratic task costs and linear correction bounds define a convex QP over $\Delta A$. The first optimized correction is applied to the base action, and refinement repeats after the next observation. Both the base policy and world model remain frozen during deployment.}
    \label{fig:framework}
    \vspace{-15pt}
\end{figure*}

\begin{algorithm}[t]
    \caption{Fast Inference-Time Policy Refinement}
    \label{alg:aawm}
    \normalfont\normalsize
    \algrenewcommand\alglinenumber[1]{\scriptsize\bfseries #1}
    \algrenewcommand\algorithmicrequire{\textbf{Input:}}
    \algrenewcommand\algorithmicindent{1em}
    \begin{algorithmic}[1]
        \Require Frozen base policy $\pi$ and frozen world model $\mathrm{A^2WM}_\theta$; prediction horizon $H$, history length $L$, and policy update interval $K$.
        \State Initialize state/action history $\mathcal H$, proposal buffer $\mathcal B$, and zero previous correction.
        \For{each control step $t=0,1,\ldots$}
        \State Observe robot/object state $x_t$; update $\mathcal H$ with the latest $L$ states and intervening executed actions.
        \Statex \hspace{\algorithmicindent}{\fontfamily{pcr}\selectfont // Base policy: every $K$ steps.}
        \If{$t\bmod K=0$}
        \State Sample a nominal action chunk from $\pi$ into $\mathcal B$.
        \EndIf
        \State Read the next $H$ commands as $A^{\mathrm{base}}$; repeat the last buffered command if needed.
        \Statex \hspace{\algorithmicindent}{\fontfamily{pcr}\selectfont // World model: one evaluation.}
        \State \textcolor{refinementblue}{$(F,\bar Z)\leftarrow\mathrm{A^2WM}_\theta(\mathcal H,A^{\mathrm{base}})$.}
        \Statex \hspace{\algorithmicindent}$\bar Z$: nominal task trajectory; $F$: action-response matrix.
        \State Set task targets $Z^{\mathrm{ref}}$, task weights $Q\succeq0$, quadratic correction penalty $\mathcal R$, and correction bounds $l,u$ from current feedback.
        \Statex \hspace{\algorithmicindent}{\fontfamily{pcr}\selectfont // Fast QP-based refinement.}
        \State \textcolor{refinementblue}{Solve for the $H$-step action correction sequence $\Delta A^\star$} \textbf{\textcolor{refinementred}{in milliseconds}} \textcolor{refinementblue}{using the modern QP solver:}
        \Statex \hspace{\algorithmicindent}\textcolor{refinementblue}{$\displaystyle
            \Delta A^\star=\arg\min_{l\leq\Delta A\leq u}
            \Bigl(\|F\Delta A+\bar Z-Z^{\mathrm{ref}}\|_Q^2+\mathcal R(\Delta A)\Bigr)$.}
        \State Apply only the first correction in $\Delta A^\star$ to the current nominal action $a_t^{\mathrm{base}}$; execute the corrected command $a_t$.
        \State Record the executed action and correction; advance $\mathcal B$ by one step.
        \EndFor
    \end{algorithmic}
\end{algorithm}

\subsection{Related Work}

One approach uses online reinforcement learning (RL) to train a residual policy that maps feedback directly to corrective actions, enabling efficient refinement at deployment~\cite{silver2018residual,johannink2018residual}. Recent work extends residual RL to precise assembly~\cite{ankile2024refinement} and pretrained imitation policies~\cite{yuan2024decorator}, while incorporating learned dynamics to inform corrective actions~\cite{gong2025korr}. Although effective, learning these residual policies requires online environment interaction, which can incur substantial data-collection costs.

Inference-time enhancement offers an alternative that does not require training a residual policy through online RL. It uses models learned offline to improve actions while keeping the pretrained policy frozen. Offline-learned value functions can rank policy samples~\cite{nakamoto2024vgps}, while guidance toward the expert state distribution can steer diffusion sampling~\cite{sun2025lpb}. World-model-based approaches use predicted action consequences to guide optimization~\cite{chua2018pets}, including manipulation through predicted images or visual features~\cite{finn2017foresight,zhou2024dinowm}. GPC brings this approach to diffusion-policy refinement by ranking sampled proposals or optimizing them through a learned world model~\cite{qi2025gpc}. Policy proposals narrow the search, but evaluating many candidates or repeatedly differentiating nonlinear predictions can still incur substantial online computation, limiting how frequently refinement can use new feedback.

Reducing this computational burden calls for world models that simplify the action optimization itself. Nonlinear model predictive control (MPC) can construct quadratic programming subproblems through local linearization, but this requires online model derivatives~\cite{diehl2005rti}. Learning models with a structure suited to control offers another route. For example, Koopman-based methods use nonlinear state lifting to construct linear predictors for control~\cite{williams2015edmd,lusch2018deep,korda2018koopman}. \textit{These approaches motivate learning predictive models whose structure simplifies the subsequent action optimization.}

Policy refinement provides a particular opportunity to exploit such structure. The pretrained policy supplies the nominal behavior, allowing the model to focus on how bounded corrections change the future trajectory near that proposal. \textit{This restricted optimization domain motivates learning a local affine response to policy corrections, without requiring globally linear physical dynamics.}

\subsection{Proposed Approach}

We propose an \textit{action-affine world model} (A\textsuperscript{2}WM) for fast inference-time refinement of frozen imitation policies (\hyperref[fig:framework]{Fig.~\ref*{fig:framework}}). Conditioned on interaction history and the current policy proposal, the model predicts a nominal trajectory and an action-response matrix. \textit{We learn this response matrix offline and predict it directly at deployment, rather than obtain it through online differentiation of a nonlinear world model.} The future trajectory is expressed as the nominal prediction plus a linear response to action corrections; both terms can depend nonlinearly on the observed context and proposal. With quadratic task costs and linear correction constraints, this structure enables refinement through a convex quadratic program (QP), avoiding candidate search and iterative nonlinear optimization.

State lifting and direct multi-step prediction support this optimization structure. We include task-relevant quantities in the prediction space so that the controller can penalize their errors directly, while learned nonlinear features support expressive prediction. Direct multi-step outputs supply the nominal trajectory and response matrix for the entire horizon without recursive model evaluations~\cite{terzi2018multistep,wu2026multistep}. We train the model offline on frozen-policy rollouts with action perturbations and keep both policy and model weights frozen at deployment. At each control step, the controller optimizes a correction sequence, applies its first correction to the base action, and repeats after new feedback. \textit{Each refinement update therefore requires only one world-model evaluation and one QP solve.}

\subsection{Contributions}

The main contributions of this work are threefold:

\begin{itemize}
    \item \textbf{Fast policy refinement.} To the best of our knowledge, we present the first framework combining an action-affine world model (A\textsuperscript{2}WM) with QP-based inference-time refinement of frozen imitation policies. Each control step only requires one model evaluation and one convex QP solve.
    \item \textbf{Approximation analysis.} We show that, at fixed context, direct affine prediction admits no larger minimum mean-squared error than recursive affine prediction, and derive a local prediction-error bound.
    \item \textbf{Empirical validation of performance and efficiency}. Across four manipulation tasks, our framework achieves the highest task-averaged success among evaluated baselines with 4.2 ms mean refinement latency per control step, excluding policy sampling. Controlled ablations assess the contributions of key design choices.
\end{itemize}

\section{Method Overview}
\label{sec:method-overview}

We formulate refinement as a finite-horizon correction of a frozen policy, then choose a prediction representation and model structure that make the online optimization a convex QP (\hyperref[fig:framework]{Fig.~\ref*{fig:framework}}). Let $x_t\in\mathbb R^{n_x}$ denote the measured robot and object state, and let $a_t^{\mathrm{base}}$ be the command proposed by the frozen policy $\pi$. A correction $\Delta a_t\in\mathbb R^m$ gives $a_t=\mathcal T(a_t^{\mathrm{base}},\Delta a_t)$, where $\mathcal T$ applies additive position corrections and composes local rotation corrections for pose commands. The history $\mathcal H=(x_{t-L+1:t},a_{t-L+1:t-1})$ contains $L$ states and the intervening executed commands. Over a horizon of $H$ steps, define $A^{\mathrm{base}}=\operatorname{col}(a_t^{\mathrm{base}},\ldots,a_{t+H-1}^{\mathrm{base}})$ and $\Delta A=\operatorname{col}(\Delta a_t,\ldots,\Delta a_{t+H-1})$, where $\operatorname{col}$ stacks its arguments vertically. We suppress the current-time index on horizon quantities for readability.

With a general one-step world model $\hat f_\theta$ parameterized by $\theta$, predictive refinement can be written in the following MPC form~\cite{mayne2000mpc}:
\begin{equation}
    \begin{alignedat}{2}
        \min_{\{\Delta a_{t+k}\}_{k=0}^{H-1}}\quad & \underbrace{\sum_{k=0}^{H-1}\ell_k(\hat x_{t+k+1})}_{\text{Task cost}}+\underbrace{\vphantom{\sum_{k=0}^{H-1}\ell_k(\hat x_{t+k+1})}\mathcal R(\Delta A)}_{\text{Regularization}} & & \\[4pt]
        \text{s.t.}\quad & \hat x_{t+k+1}=\hat f_\theta(\hat x_{t+k},a_{t+k};\mathcal H) & & \;\text{\scriptsize (Dynamics)} \\
        & a_{t+k}=\mathcal T(a_{t+k}^{\mathrm{base}},\Delta a_{t+k}) & & \;\text{\scriptsize (Corrected action)} \\
        & g_k(\hat x_{t+k+1},\Delta a_{t+k})\leq0 & & \;\text{\scriptsize (Feasibility)} \\
        & \hat x_t=x_t & & \;\text{\scriptsize (Initial state)}
    \end{alignedat}
    \label{eq:problem}
\end{equation}
The stage constraints apply for $k=0,\ldots,H-1$. Here hats denote predictions, $\ell_k$ measures task error, $\mathcal R$ penalizes correction magnitude and variation, and $g_k$ collects constraints. The history and proposal are fixed inputs to each optimization. In general, nonlinear predictions, task costs, and constraints make this problem expensive to solve repeatedly. Our design addresses the task representation first, then the dependence of predictions on corrections.

\textbf{Task-relevant state lifting.} We include the quantities needed to evaluate task errors among the model's outputs. Specifically, $z_t=\chi(x_t)\in\mathbb R^d$ retains measured states and augments them with task-relevant geometry, such as relative positions and orientations. Define $Z=\operatorname{col}(z_{t+1},\ldots,z_{t+H})$. Following lifted control formulations~\cite{korda2018koopman}, we express squared task errors and affine bounds in these coordinates as
\begin{equation}
    \begin{aligned}
        \mathcal J & = \|\widehat Z-Z^{\mathrm{ref}}\|_Q^2+\mathcal R(\Delta A), \\
        \mathcal G & = G_z\widehat Z+G_a\Delta A-g\leq0.
    \end{aligned}
    \label{eq:lifted-task}
\end{equation}
Here $\mathcal J$ is the horizon cost, $\mathcal G$ stacks the constraints, $Z^{\mathrm{ref}}$ stacks task targets, $\|v\|_Q^2=v^\top Qv$ with $Q\succeq0$, and $G_z,G_a,g$ specify linear inequalities. The targets, weights, and bounds are selected from current observations and task-stage rules, then fixed during the solve. We use the quadratic regularizer $\mathcal R(\Delta A)=\tfrac12\Delta A^\top R_q\Delta A+r_q^\top\Delta A$, with $R_q\succeq0$ and linear coefficient $r_q$, omitting constant terms. It penalizes correction magnitude and temporal variation. \textit{Lifting makes the task quantities explicit; it does not by itself make their dynamics linear.}

\textbf{Action-affine world model.} A\textsuperscript{2}WM maps the observed history and current policy proposal to the coefficients of an affine horizon prediction:
\begin{equation}
    (\mathcal H,A^{\mathrm{base}})\longmapsto(F,\bar Z),\qquad
    \widehat Z=F\Delta A+\bar Z,
    \label{eq:stacked-predictor}
\end{equation}
Here $\bar Z\in\mathbb R^{Hd}$ is the predicted trajectory at zero correction, and $F\in\mathbb R^{Hd\times Hm}$ maps corrections to changes in future task quantities. The response is causal: a future correction cannot affect an earlier predicted state. \textit{Both coefficients depend nonlinearly on the history and proposal but remain fixed during each solve, making $\widehat Z$ affine in the decision variable $\Delta A$.} The model supplies the full-horizon prediction in one evaluation; its parameterization and training are described in Sec.~\ref{sec:world-model}.

\textbf{Fast QP-based refinement.} Substituting~\eqref{eq:stacked-predictor} into~\eqref{eq:lifted-task} yields a convex QP approximation of~\eqref{eq:problem}:
\begin{equation}
    \boxed{\begin{aligned}
            \min_{\Delta A}\quad
             & \underbrace{\|F\Delta A+\bar Z-Z^{\mathrm{ref}}\|_Q^2}_{\text{Task cost}}
            +\underbrace{\vphantom{\|F\Delta A+\bar Z-Z^{\mathrm{ref}}\|_Q^2}\mathcal R(\Delta A)}_{\text{Regularization}} \\[4pt]
            \text{s.t.}\quad
             & (G_zF+G_a)\Delta A\leq g-G_z\bar Z
            \;\text{\scriptsize (Linear constraints)}
        \end{aligned}}
    \label{eq:overview-qp}
\end{equation}
Since $Q,R_q\succeq0$, the Hessian $H_q=2F^\top QF+R_q\succeq0$, establishing convexity. Our implementation uses componentwise bounds $l\leq\Delta A\leq u$, with $l,u\in\mathbb R^{Hm}$, combining correction limits and admissible command ranges (\hyperref[fig:framework]{Fig.~\ref*{fig:framework}}). HPIPM~\cite{frison2020hpipm} solves the QP using the predicted coefficients, without model rollouts or differentiation inside the solver.

The policy refreshes its action chunk every $K$ control steps. At each step, the buffer supplies the next $H$ nominal commands, repeating the last available command when needed. Using the latest feedback, we recompute $(F,\bar Z)$, solve the QP, execute only the first corrected action, and advance the buffer. \textit{Each refinement update therefore requires one world-model evaluation and one QP solve, with both policy and model weights frozen.} \hyperref[alg:aawm]{Alg.~\ref*{alg:aawm}} summarizes this procedure.

\section{Learning the Action-Affine World Model}
\label{sec:world-model}

We learn the nominal trajectory $\bar Z$ and response matrix $F$ jointly from frozen-policy rollouts. Nonlinear context encoding supplies expressive predictions, while the affine correction dependence enables the QP in Sec.~\ref{sec:method-overview}.

\subsection{Action-Affine Model Parameterization}

Given the observed history and a base-policy proposal, the model outputs a nominal trajectory $\bar Z$ and a causal action-response matrix $F$. We parameterize these outputs using nonlinear features and context encoding, augmenting physical observables with a learned dictionary~\cite{williams2015edmd,lusch2018deep}:
\begin{equation}
    \begin{aligned}
        \eta_t & =\operatorname{col}(1,\tilde z_t,\varphi_\theta(\tilde z_t)), \\
        c_i    & =\operatorname{Enc}_\theta(\mathcal H,A^{\mathrm{base},(i)}).
    \end{aligned}
    \label{eq:lift}
\end{equation}
Here $\tilde z_t$ normalizes $z_t=\chi(x_t)$, $\varphi_\theta$ is a learned dictionary, and $A^{\mathrm{base},(i)}$ contains the first $i$ proposed commands. The networks $\mathcal N_{\theta,i}$ and $\mathcal F_{\theta,ij}$ parameterize the nominal prediction and action response as
\begin{equation}
    \begin{aligned}
        \bar z_{t+i} & =z_t+\mathcal N_{\theta,i}(\eta_t,c_i,A^{\mathrm{base},(i)}),              \\
        F_{ij}       & =\mathcal F_{\theta,ij}(\eta_t,c_i,A^{\mathrm{base},(i)}),\quad 0\leq j<i, \\
        \hat z_{t+i} & =\bar z_{t+i}+\sum_{j=0}^{i-1}F_{ij}\Delta a_{t+j},
    \end{aligned}
    \label{eq:direct-predictor}
\end{equation}
for $i=1,\ldots,H$, with $F_{ij}\in\mathbb R^{d\times m}$. A causal mask sets $F_{ij}=0$ for $j\geq i$. Stacking the predictions in temporal order gives the block lower-triangular response matrix in~\eqref{eq:stacked-predictor}:
\begin{equation}
    F=
    \begin{bmatrix}
        F_{1,0} & 0       & \cdots & 0         \\
        F_{2,0} & F_{2,1} & \cdots & 0         \\
        \vdots  & \vdots  & \ddots & \vdots    \\
        F_{H,0} & F_{H,1} & \cdots & F_{H,H-1}
    \end{bmatrix}
    \in\mathbb R^{Hd\times Hm}.
    \label{eq:causal-response}
\end{equation}
Block row $i$ describes the response of $\hat z_{t+i}$ to the corrections; block column $j+1$ corresponds to $\Delta a_{t+j}$. Each zero denotes a $d\times m$ block, ensuring that a correction cannot affect a state predicted before that correction is applied. \textit{The nominal trajectory and response matrix are computed from the observed context and proposal and remain fixed during optimization.}

\subsection{Direct Multi-Step Prediction}
\label{sec:direct-model-design}

\textit{Direct multi-step prediction fits the horizon response without enforcing the cross-step factorization imposed by a recursive model.} Both formulations remain affine in action corrections and therefore compatible with QP refinement. Recursion couples future responses through shared transition matrices, whereas direct prediction learns the causal response blocks without this restriction.

To analyze this design choice, we compare our parameterization with a recursive affine predictor at a fixed initial state, history, and proposal:
\begin{equation}
    \begin{aligned}
        \hat\eta_{i+1} & =T_i\hat\eta_i+B_i\Delta a_{t+i}+d_i,   \\
        \hat z_{t+i+1} & =C\hat\eta_{i+1},\qquad i=0,\ldots,H-1.
    \end{aligned}
    \label{eq:recursive-affine-comparator}
\end{equation}
Here $\hat\eta_0=\eta_t\in\mathbb R^{n_\eta}$; $T_i$, $B_i$, and $d_i$ specify the transition, action response, and offset, while $C$ reads out physical observables. Coefficients may vary with the observed context, proposal, and horizon index, but are fixed with respect to $\Delta A$. Matrix products place larger indices on the left; empty products equal the identity. Unrolling gives $\widehat Z_{\mathrm{rec}}=\bar Z_{\mathrm{rec}}+F_{\mathrm{rec}}\Delta A$, where
\begin{equation}
    \setlength{\arraycolsep}{2pt}
    F_{\mathrm{rec}}=
    \begin{bmatrix}
        CB_0                                       & 0                                          & \cdots & 0        \\
        CT_1B_0                                    & CB_1                                       & \cdots & 0        \\
        \vdots                                     & \vdots                                     & \ddots & \vdots   \\
        C\bigl(\prod_{\ell=1}^{H-1}T_\ell\bigr)B_0 & C\bigl(\prod_{\ell=2}^{H-1}T_\ell\bigr)B_1 & \cdots & CB_{H-1}
    \end{bmatrix}.
    \label{eq:recursive-response-factorization}
\end{equation}
The zero-correction nominal trajectory is
\begin{equation}
    \setlength{\arraycolsep}{2pt}
    \bar Z_{\mathrm{rec}}
    =\begin{bmatrix}
        CT_0    \\
        CT_1T_0 \\
        \vdots  \\
        C\bigl(\prod_{\ell=0}^{H-1}T_\ell\bigr)
    \end{bmatrix}\eta_t
    +\begin{bmatrix}
        Cd_0          \\
        C(T_1d_0+d_1) \\
        \vdots        \\
        \displaystyle\sum_{j=0}^{H-1}C\bigl(\prod_{\ell=j+1}^{H-1}T_\ell\bigr)d_j
    \end{bmatrix}.
    \label{eq:recursive-nominal-trajectory}
\end{equation}
The two terms propagate the initial state and accumulate offsets. The shared transition factors constrain the horizon predictions. The following theorem shows that removing these factorization constraints cannot increase the optimal multi-step squared prediction error.

\begingroup
\setlength{\abovedisplayskip}{4pt plus 1pt minus 1pt}
\setlength{\belowdisplayskip}{4pt plus 1pt minus 1pt}
\setlength{\abovedisplayshortskip}{2pt plus 1pt}
\setlength{\belowdisplayshortskip}{3pt plus 1pt minus 1pt}
\noindent\textbf{Theorem 1 (Finite-horizon affine approximation).} At fixed context, let $\mathcal M_{\mathrm{dir}}$ contain all causal affine maps $g(\Delta A)=\bar Z+F\Delta A$, and $\mathcal M_{\mathrm{rec}}$ the maps generated by~\eqref{eq:recursive-affine-comparator} at fixed dimension $n_\eta$. For $(\Delta A,Z)$ with finite second moments, define
\begin{equation}
    \mathcal E(g)=\mathbb E\!\left[\|Z-g(\Delta A)\|_2^2\right].
    \label{eq:horizon-prediction-risk}
\end{equation}
Then
\begin{equation}
    \inf_{g\in\mathcal M_{\mathrm{dir}}}\mathcal E(g)
    \leq\inf_{g\in\mathcal M_{\mathrm{rec}}}\mathcal E(g).
    \label{eq:direct-recursive-risk}
\end{equation}
Moreover, for a minimizer $g^\star_{\mathrm{dir}}$ over $\mathcal M_{\mathrm{dir}}$ and every $g_{\mathrm{rec}}\in\mathcal M_{\mathrm{rec}}$,
\begin{equation}
    \mathcal E(g_{\mathrm{rec}})=\mathcal E(g^\star_{\mathrm{dir}})+\mathbb E\!\left[\|g_{\mathrm{rec}}(\Delta A)-g^\star_{\mathrm{dir}}(\Delta A)\|_2^2\right].
    \label{eq:affine-risk-decomposition}
\end{equation}
\noindent\textbf{Proof.} Unrolling yields $\mathcal M_{\mathrm{rec}}\subseteq\mathcal M_{\mathrm{dir}}$, proving~\eqref{eq:direct-recursive-risk}. Since $\mathcal M_{\mathrm{dir}}$ is a finite-dimensional linear subspace of square-integrable predictions, its least-squares residual is orthogonal to $g_{\mathrm{rec}}-g^\star_{\mathrm{dir}}$. Expanding the squared error gives~\eqref{eq:affine-risk-decomposition}. $\square$
\par\endgroup

\noindent\textbf{Remark.} The comparison permits full-horizon supervision for both model classes. \textit{It therefore supports direct prediction through its less restrictive parameterization, rather than through a difference in supervision horizon.}

\subsection{Training Objectives and Data Augmentation}
\label{sec:learning}

Training must fit both the nominal trajectory and its response to action corrections. The following bound separates their contributions to local prediction error.

\begingroup
\setlength{\abovedisplayskip}{4pt plus 1pt minus 1pt}
\setlength{\belowdisplayskip}{4pt plus 1pt minus 1pt}
\setlength{\abovedisplayshortskip}{2pt plus 1pt}
\setlength{\belowdisplayshortskip}{3pt plus 1pt minus 1pt}
\noindent\textbf{Theorem 2 (Local prediction error).} For a fixed initial state, proposal, and disturbance realization, let $Z^\star:\mathbb R^{Hm}\to\mathbb R^{Hd}$ be the true horizon response. Assume it is continuously differentiable with a $\kappa$-Lipschitz Jacobian $J$ on a neighborhood of $\mathbb B_r=\{v:\|v\|_2\leq r\}$, where $r>0$. Define
\[
    \epsilon_0=\|Z^\star(0)-\bar Z\|_2,\qquad
    \epsilon_F=\|J(0)-F\|_2.
\]
Then, for every $\Delta A\in\mathbb B_r$,
\begin{equation}
    \|Z^\star(\Delta A)-(\bar Z+F\Delta A)\|_2\leq\underbrace{\epsilon_0}_{\text{nominal}}+\underbrace{\epsilon_F\|\Delta A\|_2}_{\text{response}}+\underbrace{\tfrac{\kappa}{2}\|\Delta A\|_2^2}_{\text{curvature}}.
    \label{eq:local-error}
\end{equation}
\noindent\textbf{Proof.} Taylor's integral remainder gives
\[
    \begin{aligned}
        Z^\star(\Delta A) & =Z^\star(0)+J(0)\Delta A+r_Z,            \\
        \|r_Z\|_2         & \leq\int_0^1\kappa s\|\Delta A\|_2^2\,ds
        =\tfrac{\kappa}{2}\|\Delta A\|_2^2.
    \end{aligned}
\]
Subtracting $\bar Z+F\Delta A$ and applying the triangle inequality proves~\eqref{eq:local-error}. $\square$
\par\endgroup

Within a smooth local regime, the bound separates \textit{nominal error, response error, and curvature}. Correction bounds limit the contribution of the latter two terms, motivating a model trained for accurate prediction within this neighborhood. We use perturbed rollouts to jointly supervise the nominal forecast and action response over the full horizon, and recentering to provide an explicit zero-correction target for the nominal forecast.

\begin{figure}[t]
    \centering
    \includegraphics[width=\columnwidth,pagebox=cropbox,clip]{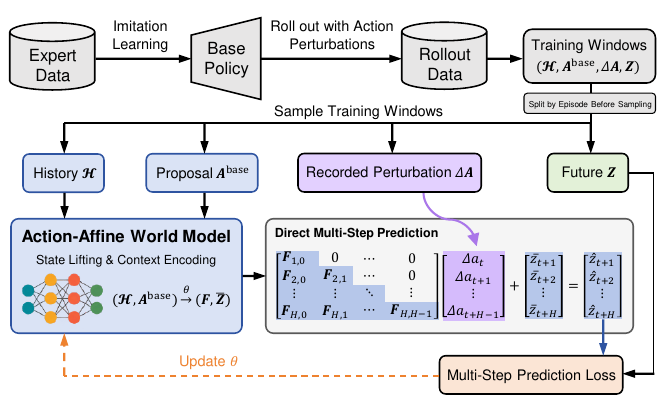}
    \caption{\textbf{Offline world-model learning.} Frozen-policy rollouts with action perturbations provide windows $(\mathcal H,A^{\mathrm{base}},\Delta A,Z)$. Direct multi-step supervision trains $\widehat Z=F\Delta A+\bar Z$; recentering reuses executed trajectories to supervise the nominal forecast. Episodes are split before sampling windows.}
    \label{fig:world-model-training}
    \vspace{-15pt}
\end{figure}

Perturbed policy rollouts provide excitation for learning action responses~\cite{ljung1999identification}. Each window $(\mathcal H,A^{\mathrm{base}},\Delta A,Z)$ records the proposal available at its start and corrections that recover the executed commands, including subsequent proposal refreshes (\hyperref[fig:world-model-training]{Fig.~\ref*{fig:world-model-training}}). We supervise the full horizon with
\begin{equation}
    \mathcal L_{\rm pred}=\mathbb E\!\left[\frac1{Hd}\sum_{i=1}^{H}\sum_{p=1}^{d}\rho\!\left(\frac{\hat z_{t+i,p}-z_{t+i,p}}{\sigma_{i,p}}\right)\right],
    \label{eq:training}
\end{equation}
where $\rho$ is Smooth L1, $p$ indexes observables, and $\sigma_{i,p}>0$ is the training-set increment standard deviation in the prediction coordinates, using unit scales for constant channels. Expectations average training windows.

\textit{Recentering directly supervises the nominal forecast without additional interaction.} Relabel the same window as $(\mathcal H,A,0,Z)$, where $A=\operatorname{col}(a_t,\ldots,a_{t+H-1})$ is the executed sequence, and recompute the proposal context. The recentering loss $\mathcal L_{\rm ctr}$ uses~\eqref{eq:training} with $\hat z_{t+i,p}$ replaced by $\bar z_{t+i,p}(\mathcal H,A)$, retaining the same targets and normalization. This provides a measured zero-correction target at the executed proposal without counterfactual rollouts.

We train jointly offline with $\mathcal L=\mathcal L_{\rm pred}+\mathcal L_{\rm ctr}$. At deployment, the frozen model maps the latest history and proposal to $(F,\bar Z)$ in one evaluation, supplying the action-affine prediction needed for QP refinement without online model differentiation.

\begin{table*}[t]
    \centering
    \caption{Success rate (mean $\pm$ std, \%) and mean refinement latency per control step (ms).}
    \label{tab:success-refinement}
    \footnotesize
    \setlength{\tabcolsep}{1.2pt}
    \renewcommand{\arraystretch}{1.15}
    \begin{tabular}{lcccccccccc}
        \toprule
                      & \multicolumn{2}{c}{OneLeg} & \multicolumn{2}{c}{RoundTable} & \multicolumn{2}{c}{Lamp} & \multicolumn{2}{c}{PushT} & \multicolumn{2}{c}{Mean}                                                                                                                          \\
        \cmidrule(lr){2-3}\cmidrule(lr){4-5}\cmidrule(lr){6-7}\cmidrule(lr){8-9}\cmidrule(l){10-11}
        Method        & Success Rate $\uparrow$    & Latency $\downarrow$           & Success Rate $\uparrow$  & Latency $\downarrow$      & Success Rate $\uparrow$  & Latency $\downarrow$ & Success Rate $\uparrow$ & Latency $\downarrow$ & Success Rate $\uparrow$ & Latency $\downarrow$ \\
        \midrule
        DP            & $53.13\pm2.71$             & --                             & $6.25\pm1.56$            & --                        & $6.77\pm3.25$            & --                   & $42.19\pm5.63$          & --                   & $27.08\pm1.97$          & --                   \\
        LPB           & $53.65\pm2.39$             & 1.10                           & $5.73\pm3.25$            & 3.58                      & $3.13\pm1.56$            & 1.35                 & $42.71\pm7.86$          & 5.12                 & $26.30\pm0.81$          & 2.79                 \\
        GPC-RANK      & $58.85\pm7.71$             & 8.46                           & $2.60\pm2.39$            & 16.10                     & $8.85\pm2.39$            & 8.75                 & $\mathbf{73.96}\pm7.38$ & 67.75                & $36.07\pm1.63$          & 25.27                \\
        GPC-OPT       & $\mathbf{75.52}\pm3.25$    & 449.90                         & $8.33\pm2.39$            & 432.58                    & $29.17\pm7.71$           & 440.79               & $69.27\pm3.93$          & 432.23               & $45.57\pm2.15$          & 438.88               \\
        \rowcolor{oursrow}
        \textbf{Ours} & $\mathbf{75.52}\pm5.92$    & 3.67                           & $\mathbf{42.71}\pm1.80$  & 3.79                      & $\mathbf{35.94}\pm12.20$ & 4.04                 & $\mathbf{73.96}\pm5.02$ & 5.22                 & $\mathbf{57.03}\pm3.05$ & 4.18                 \\
        \bottomrule
    \end{tabular}
    \vspace{-10pt}
\end{table*}

\section{Experiments}
\label{sec:experiments}

We evaluate policy refinement from four perspectives: task performance, online computation, prediction structure, and learning mechanisms. Our experiments address four research questions:
\begin{itemize}
    \item \textbf{RQ1:} How effectively does A\textsuperscript{2}WM refine frozen imitation policies across manipulation tasks?
    \item \textbf{RQ2:} How much online computation does A\textsuperscript{2}WM require, and where do its computational savings come from?
    \item \textbf{RQ3:} How does the world model structure affect prediction accuracy and control performance?
    \item \textbf{RQ4:} How do the key components of our framework contribute to its performance?
\end{itemize}

\subsection{Experimental Setup}
\label{sec:setup}

\begin{figure}[t]
    \centering
    \includegraphics[width=\columnwidth,pagebox=cropbox,clip]{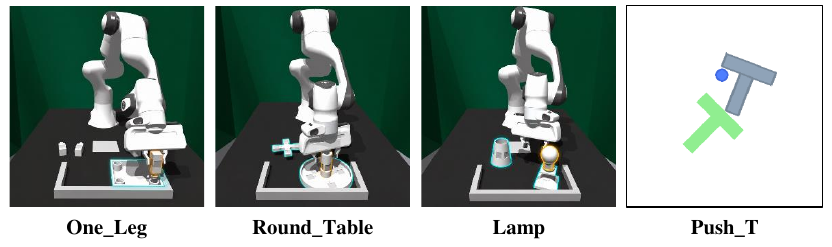}
    \caption{\textbf{Task visualizations.} Snapshots of the \textit{FurnitureBench} assembly tasks and the planar PushT task.}
    \label{fig:task-visualization}
    \vspace{-15pt}
\end{figure}

\textbf{Benchmarks and Baselines.} We evaluate on the FurnitureBench~\cite{heo2023furniturebench} tasks OneLeg, RoundTable, and Lamp in Isaac Gym~\cite{makoviychuk2021isaacgym}, and on planar PushT~\cite{chi2023diffusion} (\hyperref[fig:task-visualization]{Fig.~\ref*{fig:task-visualization}}). We compare against the following baselines: (1) \textbf{DP}~\cite{chi2023diffusion}, which executes action chunks from the frozen diffusion policy without refinement; (2) \textbf{LPB}~\cite{sun2025lpb}, which guides denoising toward expert features; (3) \textbf{GPC-RANK}, which selects among 100 proposals; and (4) \textbf{GPC-OPT}, which refines a proposal through 24 gradient-based optimization steps~\cite{qi2025gpc}. The GPC methods share a recursive nonlinear world model trained with multi-step supervision. Within each task, all methods use the same DP checkpoint; A\textsuperscript{2}WM and GPC share dynamics data, episode splits, task objectives, correction bounds, and an eight-step prediction horizon. LPB retains its expert-distribution objective and eight-step feedback. All reported quantitative comparisons use the standard controller configuration.

\textbf{Tasks and Metrics.} For the furniture tasks, success is determined by the benchmark's assembly criteria. For PushT, success requires more than 95\% coverage of the target region. Each task--method pair is evaluated on three seeds with 64 episodes each. Timing uses three sequential single-environment episodes per task and method on the same RTX 4090 workstation, with neural-network inference on the GPU and QP solving on the CPU. We distinguish total inference time, which includes DP sampling, from refinement latency, which excludes it. For GPC-RANK, we subtract Ours' single-proposal DP time to retain the cost of additional candidates. Task averages give each task equal weight.

\begin{table}[t]
    \centering
    \caption{Learning paradigm and additional interaction data per task.}
    \label{tab:data-requirements}
    \footnotesize
    \setlength{\tabcolsep}{3pt}
    \renewcommand{\arraystretch}{1.15}
    \begin{tabular}{p{0.2\columnwidth}p{0.3\columnwidth}p{0.4\columnwidth}}
        \toprule
        Method                            & On-Policy Interaction & Data for Refinement                      \\
        \midrule
        ResiP~\cite{ankile2024refinement} & Yes                   & $\approx5\times10^5$ episode-equivalents \\
        KoRR~\cite{gong2025korr}          & Yes                   & $\approx1\times10^5$ episode-equivalents \\
        \rowcolor{oursrow}
        A\textsuperscript{2}WM (ours)     & No                    & 256--512 episodes                        \\
        \bottomrule
    \end{tabular}
\end{table}

\subsection{Policy Refinement Performance}
\label{sec:performance}

To answer \textbf{RQ1}, \hyperref[tab:success-refinement]{Tab.~\ref*{tab:success-refinement}} and \hyperref[fig:success-latency]{Fig.~\ref*{fig:success-latency}(a)} compare task success. A\textsuperscript{2}WM raises mean success from 27.08\% for DP to 57.03\%, achieving the highest observed task average among the evaluated methods. The largest gains over the refinement baselines occur on RoundTable and Lamp, where Ours reaches 42.71\% and 35.94\% despite DP succeeding in fewer than 7\% of episodes. On OneLeg and PushT, Ours ties the best baseline rather than providing a further gain. \textit{A low task-completion rate does not make the base policy's proposals unusable: feedback correction can turn these proposals into substantially more successful executions.}

\textbf{Data required for refinement.}\label{sec:data-requirements} These gains in task success are obtained using fixed offline datasets of 256--512 rollout episodes per task. The reported on-policy refinement budgets in \hyperref[tab:data-requirements]{Tab.~\ref*{tab:data-requirements}} involve substantially more environment interaction. \textit{Our results show that learning action consequences from a modest offline rollout dataset can support effective feedback correction without repeatedly collecting data to update a residual policy.} This makes model-based refinement a practical option when additional environment interaction is limited.

\subsection{Computational Efficiency}
\label{sec:efficiency}

For \textbf{RQ2}, A\textsuperscript{2}WM averages 4.18\,ms of refinement per control step, approximately $6\times$ faster than GPC-RANK and $105\times$ faster than GPC-OPT (\hyperref[tab:success-refinement]{Tab.~\ref*{tab:success-refinement}}). Including DP sampling, inference time increases from 17.71 to 21.81\,ms (\hyperref[fig:success-latency]{Fig.~\ref*{fig:success-latency}(b)}). LPB incurs less overhead but does not improve mean success.

\begin{figure}[t]
    \centering
    \includegraphics[width=\columnwidth]{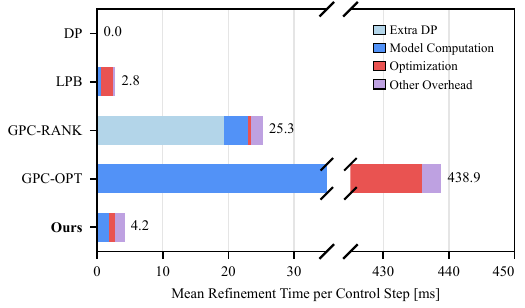}
    \caption{\textbf{Breakdown of refinement latency.} Each method's DP time is excluded; for GPC-RANK, subtracting Ours' single-proposal DP time retains excess candidate sampling as Extra DP. Model Computation covers prediction; Optimization covers candidate scoring, gradient-based updates, or QP solving. Other Overhead contains the remaining online work. The near-zero DP entry is residual overhead, not policy-sampling time.}
    \label{fig:inference-breakdown}
    \vspace{-15pt}
\end{figure}

\hyperref[fig:inference-breakdown]{Fig.~\ref*{fig:inference-breakdown}} identifies the bottlenecks. GPC-RANK spends 19.39\,ms on additional proposal sampling, accounting for 77\% of refinement latency, versus 3.63\,ms on batched prediction and 0.39\,ms on scoring and selection. Candidate generation therefore dominates despite parallel evaluation.

For GPC-OPT, each action update requires a fresh eight-step nonlinear rollout and backpropagation. Its 24 updates plus a final evaluation accumulate 185.17\,ms of model computation and 250.73\,ms of optimization work, including scoring, backpropagation, and updates---over 99\% of refinement latency. \textit{Repeated prediction and differentiation, rather than the update arithmetic, make this online optimization expensive.}

A\textsuperscript{2}WM computes the nominal trajectory and response matrix once in 1.77\,ms, then holds them fixed during QP solution. QP processing takes 1.00\,ms; the remaining refinement time covers objective construction and other online work. \textit{Solver iterations require no further neural-network evaluation or differentiation: action optimization proceeds entirely within the convex QP.}

\subsection{World Model Structure}
\label{sec:world-model-analysis}
\label{sec:direct-prediction}

To answer \textbf{RQ3}, we vary two design choices: direct versus recursive prediction, and affine versus nonlinear action dependence. The resulting four models share datasets, multi-step supervision, approximately matched parameter budgets, and the same QP formulation. Affine models supply the action-response matrix explicitly; nonlinear models require an online Jacobian. We examine prediction accuracy first, then the cost of constructing the QP model and the resulting control performance.

\begin{figure}[t]
    \centering
    \includegraphics[width=\columnwidth]{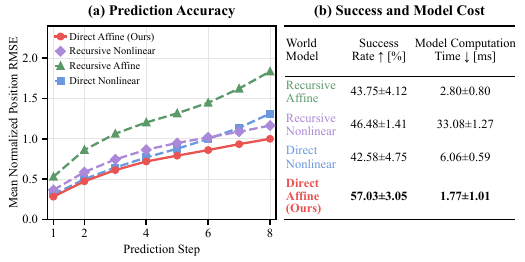}
    \caption{(a) \textbf{Effect of model structure on prediction accuracy.} Position RMSE is normalized by Ours' eighth-step RMSE within each task and averaged across four tasks. (b) \textbf{Effect of model structure on success and computation.} Success rate and model-computation time are averaged equally across tasks under the same QP formulation. Model time includes online Jacobian computation for nonlinear models but excludes DP sampling and QP processing.}
    \label{fig:world-model-analysis}
    \vspace{-15pt}
\end{figure}

\textbf{Prediction accuracy.} \hyperref[fig:world-model-analysis]{Fig.~\ref*{fig:world-model-analysis}(a)} compares position RMSE over 8192 validation windows per task, normalized by Ours' eighth-step RMSE before averaging across tasks. Direct prediction reduces the affine model's eighth-step error from 1.83 to 1.00. The nonlinear models show a smaller, opposite difference: Recursive Nonlinear attains 1.16 versus 1.31 for Direct Nonlinear. A plausible explanation is that an expressive nonlinear transition model, composed across time and trained with multi-step supervision, can also capture complex horizon responses. Thus, the accuracy benefit of direct prediction is evident within the affine models, but does not extend uniformly to the nonlinear models.

\begin{figure}[t]
    \centering
    \includegraphics[width=\columnwidth]{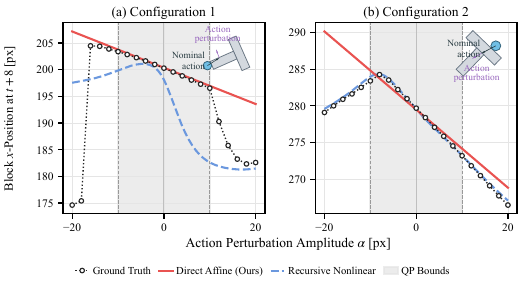}
    \caption{\textbf{Local action-response prediction in PushT.} Each panel fixes the initial state, history, and proposal, then scales a fixed action-perturbation sequence by $\alpha$. Curves show the block's $x$-position at step eight.}
    \label{fig:local-action-response}
    \vspace{-15pt}
\end{figure}

\hyperref[fig:local-action-response]{Fig.~\ref*{fig:local-action-response}} illustrates that locally near-linear action responses can occur even in contact-rich manipulation. Refinement need not capture the full nonlinear response to arbitrary actions; an affine approximation within the bounded correction neighborhood can provide useful directions for optimization.

\textbf{Computational effect of direct prediction.} We isolate model-computation time, including online Jacobian construction when required but excluding policy sampling and QP processing. Direct prediction reduces model-computation time in both model classes (\hyperref[fig:world-model-analysis]{Fig.~\ref*{fig:world-model-analysis}(b)}): from 2.80 to 1.77\,ms for affine models and from 33.08 to 6.06\,ms for nonlinear models. Recursive nonlinear prediction requires sequential model calls and derivative propagation through the predicted-state chain; direct prediction evaluates and differentiates a single horizon mapping. \textit{Direct prediction therefore offers a substantial computational benefit.}

\textbf{Control performance and computation.} Direct Nonlinear removes recursive prediction but still requires online differentiation. Holding direct prediction fixed, the explicit affine response further reduces model-computation time from 6.06 to 1.77\,ms, while mean success increases from 42.58\% to 57.03\%. Relative to Recursive Affine, Ours also improves success from 43.75\% to 57.03\% (\hyperref[fig:world-model-analysis]{Fig.~\ref*{fig:world-model-analysis}(b)}). \textit{The two design choices serve complementary roles: direct prediction avoids recursive horizon computation, while the explicit response eliminates online neural-network differentiation.} Their combination achieves the highest mean success and lowest model-computation time among these four models.

\subsection{Action-Response and Training Ablations}
\label{sec:mechanism-ablation}

For \textbf{RQ4}, we move from comparisons between independently trained model structures to mechanisms within A\textsuperscript{2}WM. We replace the predicted response matrix with the same model's nominal-head Jacobian, holding the checkpoint and nominal prediction fixed, and separately retrain without recentering. \hyperref[tab:mechanism-ablation]{Tab.~\ref*{tab:mechanism-ablation}} reports closed-loop performance across all four tasks; \hyperref[tab:mechanism-prediction]{Tab.~\ref*{tab:mechanism-prediction}} provides prediction diagnostics in PushT. At step eight, Nominal RMSE measures block-position error without corrections; Response RMSE measures error in the correction-induced position change, using paired rollouts from the same initial state.

\begin{table*}[t]
    \centering
    \caption{Effects of explicit response prediction and recentering on control performance.}
    \label{tab:mechanism-ablation}
    \footnotesize
    \setlength{\tabcolsep}{2pt}
    \renewcommand{\arraystretch}{1.15}
    \begin{tabular}{@{}lcccccc@{}}
        \toprule
                             & \multicolumn{5}{c}{Success Rate $\uparrow$ [\%]} & Mean Refinement                                                                                                                   \\
        \cmidrule(lr){2-6}
        Variant              & OneLeg                                           & RoundTable              & Lamp                    & PushT                   & Mean                    & Latency $\downarrow$ [ms] \\
        \midrule
        \rowcolor{oursrow}
        Ours                 & $75.52\pm5.92$                                   & $\mathbf{42.71\pm1.80}$ & $35.94\pm12.20$         & $73.96\pm5.02$          & $\mathbf{57.03\pm3.05}$ & $4.18\pm0.64$             \\
        Ours w/ Jacobian     & $\mathbf{82.81\pm3.12}$                          & $30.21\pm2.39$          & $30.73\pm3.93$          & $\mathbf{80.21\pm1.80}$ & $55.99\pm0.23$          & $18.85\pm6.89$            \\
        Ours w/o recentering & $69.79\pm5.02$                                   & $42.19\pm4.13$          & $\mathbf{37.50\pm2.71}$ & $70.83\pm7.05$          & $55.08\pm3.10$          & $\mathbf{4.16\pm0.61}$    \\
        \bottomrule
    \end{tabular}
\end{table*}

\textbf{Explicit action response.} Predicting $F$ explicitly gives similar mean success (57.03\% versus 55.99\%) while reducing refinement latency from 18.85 to 4.18\,ms. In the response tests, the explicit response has slightly higher RMSE (0.62 versus 0.54\,px). \textit{Its main benefit is therefore obtaining useful correction directions at substantially lower computational cost, without requiring a more accurate response than the nominal-head Jacobian.}

\begin{table}[t]
    \centering
    \caption{Effects of explicit response prediction and recentering on prediction accuracy and cost in PushT.}
    \label{tab:mechanism-prediction}
    \footnotesize
    \setlength{\tabcolsep}{3pt}
    \renewcommand{\arraystretch}{1.15}
    \begin{tabular}{@{}lccc@{}}
        \toprule
        Variant              & \shortstack{Nominal                                            \\RMSE [px]} & \shortstack{Response\\RMSE [px]} & \shortstack{Model Time\\{[ms]}} \\
        \midrule
        \rowcolor{oursrow}
        \textbf{Ours}        & $\mathbf{1.06}$     & $0.62$          & $3.51\pm0.06$          \\
        Ours w/ Jacobian     & $\mathbf{1.06}$     & $\mathbf{0.54}$ & $27.79\pm1.23$         \\
        Ours w/o recentering & $1.18$              & $0.63$          & $\mathbf{3.46\pm0.06}$ \\
        \bottomrule
    \end{tabular}
    \vspace{-10pt}
\end{table}

\textbf{Recentering.} Retraining without recentering lowers mean success from 57.03\% to 55.08\%. In the response tests, recentering reduces nominal RMSE from 1.18 to 1.06\,px, while response RMSE changes little (0.63 to 0.62\,px). This distinction matches the loss design: recentering directly supervises the nominal forecast at the executed proposal. \textit{These results suggest that recentering primarily improves nominal trajectory prediction, with a modest increase in mean task success.}

\section{Conclusion}

We presented A\textsuperscript{2}WM for fast inference-time refinement of frozen imitation policies. The model directly supplies a nominal trajectory and an action-response matrix, enabling refinement through one model evaluation and one convex QP solve per control step. Across four simulated manipulation tasks, our framework achieves the highest task-averaged success rate among the evaluated methods, with 4.18 ms of mean refinement latency per control step, excluding policy sampling. Ablations show complementary computational benefits from direct prediction and explicit action-response prediction. These results highlight the value of learning world models whose predictive structure supports efficient action optimization.

Our current implementation relies on measured robot and object states. Future work will investigate visual action-affine world models for fast online refinement without explicit robot and object state measurements.

\bibliographystyle{IEEEtran}

\bibliography{ref}

\end{document}